\documentclass[acmsmall,screen,nonacm,review=false,timestamp=false]{acmart}
\renewcommand\footnotetextcopyrightpermission[1]{} 
\usepackage[plain]{fancyref}
\usepackage[draft=true]{minted}

\usepackage{lipsum}
\usepackage{multirow}
\AtBeginDocument{%
  \providecommand\BibTeX{{%
    \normalfont B\kern-0.5em{\scshape i\kern-0.25em b}\kern-0.8em\TeX}}}

\begin{document}

\title[Generalization in Healthcare AI]{Generalization in Healthcare AI: Evaluation of a Clinical Large Language Model}


\author{Salman Rahman}
\affiliation{%
  \institution{New York University}
      \country{USA}
  }
\email{salman@nyu.edu}

\author{Lavender Yao Jiang}
\affiliation{%
  \institution{New York University}
      \country{USA}
}
\email{lyj2002@nyu.edu}


\author{Saadia Gabriel}
\affiliation{%
  \institution{New York University}
      \country{USA}
}
\email{sg8390@nyu.edu}


\author{Yindalon Aphinyanaphongs}
\affiliation{%
  \institution{NYU Langone Health}
    \country{USA}
  }
\email{Yin.A@nyulangone.org}

\author{Eric Karl Oermann}
\affiliation{%
  \institution{NYU Langone Health}
    \country{USA}
  }
\email{Eric.Oermann@nyulangone.org}

\author{Rumi Chunara}
\affiliation{%
  \institution{New York University}
  \country{USA}}
\email{rumi.chunara@nyu.edu}

\renewcommand{\shortauthors}{Rahman et al.}

\begin{abstract}

Advances in large language models (LLMs) provide new opportunities in healthcare for improved patient care, clinical decision-making, and enhancement of physician and administrator workflows. However, the potential of these models importantly depends on their ability to generalize effectively across clinical environments and populations, a challenge often underestimated in early development. To better understand reasons for these challenges and inform mitigation approaches, we evaluated ClinicLLM, an LLM trained on [HOSPITAL]'s clinical notes, analyzing its performance on 30-day all-cause readmission prediction focusing on variability across hospitals and patient characteristics. We found poorer generalization particularly in hospitals with fewer samples, among patients with government and unspecified insurance, the elderly, and those with high comorbidities. To understand reasons for lack of generalization, we investigated sample sizes for fine-tuning, note content (number of words per note), patient characteristics (comorbidity level, age, insurance type, borough), and health system aspects (hospital, all-cause 30-day readmission, and mortality rates). We used descriptive statistics and supervised classification to identify features. We found that, along with sample size, patient age, number of comorbidities, and the number of words in notes are all important factors related to generalization. Finally, we compared local fine-tuning (hospital specific), instance-based augmented fine-tuning and cluster-based fine-tuning for improving generalization. Among these, local fine-tuning proved most effective, increasing AUC by 0.25\% to 11.74\% (most helpful in settings with limited data). Overall, this study provides new insights for enhancing the deployment of large language models in the societally important domain of healthcare, and improving their performance for broader populations. ([HOSPITAL] name redacted for blind review).
\end{abstract}





\maketitle

\section{Introduction}

 Pretrained large language models (LLMs), trained on massive, diverse datasets, have been applied to downstream tasks such as multitask learning, language translation, and sentiment analysis \cite{brown2020language, Devlin2019BERTPO}. This signifies a noteworthy transformation in machine learning, emphasizing the utilization of a single pretrained model for diverse applications \cite{bommasani2021opportunities}. In healthcare, models like Med-PaLM, bioclinicalBERT, and GatorTron, have been developed to focus on applications such as medical chat-bots for clinicians and patients, interactive note-taking, bedside decision support, patient data analysis, and the generation of detailed radiology reports \cite{moor2023foundation}. These language models are trained on a combination of clinical data and non-medical, general-purpose datasets, including standard internet corpora, book, PubMed, and Wikipedia \cite{singhal2022large, waisberg2023gpt}. More specific to in-hospital tasks, \citet{jiang2023health} demonstrated an LLM trained solely on clinical data achieved higher AUC in predicting readmission tasks compared to those trained on a mix of clinical and non-clinical data (internet corpora, book, wikipedia, pubmed) \cite{yang2022large}. This study also suggests LLMs trained entirely on clinical notes can contribute to workflow improvements, cost reductions, and reduced physician burnout, by effectively predicting downstream tasks like readmission, mortality, length of stay, comorbidity, and insurance denials. Though the study made a pioneering effort demonstrating the effectiveness of training a domain specific LLM using only in house clinical notes \cite{jiang2023health}, a systematic study of how well a model generalizes across different patients groups and hospitals has not yet been studied. The consistent performance (generalization) of large language models across different situations, hospitals, and patient groups is an important challenge \cite{finlayson2021clinician, ghassemi2020review, behar2023generalization}.

The generalizability of large language models (LLMs) refers to their ability to perform consistently well across diverse situations such as different healthcare settings, diverse patient populations, and varied types of clinical notes \cite{burns2023weak}.  The generalizability issue in healthcare is particularly critical because the general failure modes of machine learning models, such as bad or inadequate data, differences or shifts in environment, and poor reporting, can have a significant impact on human health \cite{saria2019tutorial, subbaswamy2020development, futoma2020myth, singh2021fairness, singh2022generalizability}. This aspect is crucial for successfully deploying models since we want model predictions to be accurate when applied to new situations \cite{yang2022machine, subbaswamy2021evaluating}. Research efforts to enhance the generalization of LLMs have focused on increasing the amount of pre-training data, expanding model size, and incorporating auxiliary data via multitask learning and adversarial training \cite{tu2020empirical, hendrycks2019using, dong2021should, hendrycks2020pretrained}.

Accordingly, to study the generalizability of LLMs trained on clinical notes, in this research, we focus on the generalization capability of a clinical large language model ClinicLLM, which is trained exclusively on clinical notes from [HOSPITAL]. ClinicLLM is based on a state-of-the-art (SOTA) model for the 30-day all-cause readmission task, achieving an 79.9\% Area Under the Curve (AUC) surpassing baselines as demonstrated in previous work \cite{jiang2023health}. As well, in using this model we have access to system-specific data (e.g. linked patient characteristics) that we would not have access from other competitive clinical models in order to perform the analysis herein. Further, our hospital system is amenable studying generalization due to the diversity of multiple hospitals in the [HOSPITAL] health system, the diverse patient groups with different socioeconomic and health conditions, and the complexity of the clinical notes available in each hospital where the model is trained, making it a suitable choice for this study. We study this via the following three research questions:

\begin{itemize}

    \item Generalizability assessment: we analyze the performance of ClinicLLM, a Large Language Model (LLM) trained on data from four hospitals within [HOSPITAL] system, for predicting 30-day all-cause readmission by hospital and across patient groups categorized by insurance type, races, age group, and comorbidity level.

    \item Investigate reasons for lack of generalization: we investigate reasons for varying model performance, focusing on sample size for fine-tuning, and note, patient/illness and provider/system aspects by using descriptive statistics and supervised learning to identify the joint features that contribute to generalizability.

    \item Strategies for improving generalization: we test local hospital-specific fine-tuning, instance-based augmented fine-tuning (data augmentation using similar notes from other hospitals) and cluster-based fine-tuning approaches to improve performance across healthcare settings and patient demographics. 
\end{itemize}

\section{Related Work}

\subsection{Generalization in Clinical Large Language Models (LLMs)}

Research on clinical Large Language Models (LLMs) highlights their state-of-the-art performance and potential in handling medical tasks such as predicting readmission, mortality, medical question-answering, and diagnostic assistance \cite{singhal2022large, nori2023can, huang2019clinicalbert}. However, disparities have been observed in such models across racial groups (performing worse for Black patients), clinical departments (performing worst in the internal medicine department), and age demographics (worse for patients over 80 years), particularly when trained exclusively on data from a single hospital \cite{jiang2023health}. \citet{saria2019tutorial} outline four primary sources of failures in healthcare: 1) inadequate or poor-quality data, 2) environmental changes or shifts, 3) errors associated with the model itself, and 4) issues with reporting accuracy. Additionally, \citet{futoma2020myth} outline potential threats to generalizability in healthcare machine learning, including changes in practice patterns, variations between health systems, patient demographics, and genotypic/phenotypic differences, as well as hardware/software variations and diverse environmental, social, and cultural factors. Although there have been lots of progress regarding ensuring the generalizability of machine learning in healthcare, for clinical large language models there is a lack of studies exploring generalization capability, especially across different hospitals within the same system and across patients groups.

\subsection{Challenges to Generalization in Large Language Models}

LLMs encounter generalization challenges such as struggles with out-of-distribution data, vulnerability to adversarial attacks, and establishing spurious correlations \cite{hendrycks2019using}. While pre-training on diverse datasets improves generalization, especially from minority examples in training sets that diverge from common spurious patterns \cite{dong2021should}, the effectiveness of pre-training in enhancing generalization is limited in scenarios with scarce counterexamples. This limitation suggests that expanding datasets alone may not suffice, and further improvements in LLM generalization should be studied \cite{tu2020empirical}. 

\subsection{Improving Generalization in Large Language Models}

Significant efforts have been made to improve the generalization of LLMs through data augmentation and scaling of models. \citet{tu2020empirical} showed that expanding and diversifying training datasets to make models more adaptable to varied data and increasing the size and complexity of the model architectures can potentially lead to improved generalization. \citet{hendrycks2020pretrained} highlight strategies for improving generalization by better pre-training of models, including the use of more comprehensive and diverse training data, and techniques like multitask learning and domain-specific training. However, pre-training different models poses specific challenges, often related to costs, computational resources, and time. Here, we utilize a single pre-trained model, which has been trained on clinical notes from the [HOSPITAL] Electronic Health Record (EHR) system. We then explore avenues for enhancing generalization from a fine-tuning perspective, specifically focusing on fine-tuning with local, augmented, and clustered samples.

\section{Methods: Pretraining, Fine-Tuning, Implementation, Clustering and Analysis Details}
\label{sec:implementation-details}

\subsection{Clinical Prediction Task}

Predicting 30-day all-cause readmission is important as hospitals face financial penalties for high readmission rates, suggesting inadequate care or premature discharge \cite{caruana2015intelligible}. Identifying patients needing more post-discharge support can lower readmission rates, enhance hospital performance and patient care, and reduce healthcare costs \cite{davis2022effective}. This task predicts the likelihood of the patient in a specific note being readmitted within 30 days post-discharge. 

For 30-day readmission tasks, History and Physical (H\&P) notes were selected because they are available prior to other types of clinical notes, like discharge notes, thus allowing for earlier assessment of the likelihood of readmission. The notes originate from four hospitals (Hospital 1, Hospital 2 , Hospital 3, and Hospital 4 [Names redacted for blind review]), encompassing patients from five New York City boroughs (Manhattan, Queens, Bronx, Brooklyn, Staten Island). Patients at these hospitals provide representation across a mix of age groups, races, different levels of comorbidity, and insurance types. The prediction is made at note-level, which includes notes of single patients with multiple encounters in the testing dataset, and prediction is made for each note separately. IRB approval for this study was obtained from our institution.

\subsection{Pretraining}
\label{subsec: pretraining}
Pretraining of our model utilized weights from \citet{jiang2023health}, which employed a BERT base architecture with 109 million parameters using masked language modeling (MLM). The model used a dataset of clinical notes in CSV format sourced from the [HOSPITAL] Electronic Health Record (EHR) system. The dataset included records from 387,144 patients with a total of 7,247,694 notes, amounting to approximately 4.1 billion words. A large portion of the data was from Hospital 1 (records of 256,217 patients, 4,342,602 notes, and over 2.3 billion words). Another significant portion was from Hospital 2 (104,521 patients, 1,337,352 notes, and over 1.1 billion words). These notes were generated by many types of medical professionals, including physicians, residents, physician assistants, nurse practitioners, and fellows at all four hospitals, from 2011 to 2020. Further details on the model architecture and methods are available in the work by \citet{jiang2023health}.

The pretraining dataset includes nine types of clinical notes: progress notes, consultations, history and physical examination (H\&P) notes, discharge summaries, ERS update notes, operative (OP) notes, brief operative notes, emergency department (ED) provider notes, and miscellaneous notes. To prepare the dataset for pretraining, notes related to billing and those flagged as invalid or empty were excluded. The remaining data was divided into training, validation, and test sets following a 949:50:1 ratio. Consistent with BERT's methodology, 15\% of the tokens in the dataset were randomly masked \cite{Devlin2019BERTPO}. The pretrained model is hereafter referred to as ``ClinicLLM''.

\subsection{Fine-tuning}
\label{subsec: fine-tuning}

A dataset was compiled from the [HOSPITAL] Electronic Health Record (EHR), consisting of History and Physical (H\&P) notes with binary labels for readmission, intended for fine-tuning. The dataset includes 170,191 patients, 222,824 notes, and 0.27 billion words, covering encounters from December 2012 to December 2021. The data was divided into training, validation, random test, and temporal test sets. The first three sets comprise notes from December 2012 to June 2021, randomly selected with an 80-10-10 (train-val-test) ratio within this period. The temporal test set contained H\&P notes from new patients from July 2021 to December 2021, for no timeline overlap with the training data. 

The fine-tuning phase involves adding a classification layer on top of the pre-trained BERT architecture and conducting end-to-end training with a labeled 30-day readmission dataset. The maximum input sequence length is limited to 512 tokens, following the original BERT methodology \cite{Devlin2019BERTPO}. For global fine-tuning of the model on the 30-day all-cause readmission prediction task, we conducted a hyperparameter search using the Optuna framework \cite{akiba2019optuna}. This search identified an optimal learning rate of $8.37 \times 10^{-6}$ and a training batch size of 32 per device. The model underwent fine-tuning over 9 epochs, adapting it to the specific requirements of the readmission prediction task. We used the same hyperparameters for local, instance-based augmented, and cluster-based fine-tuning as those used for the global fine-tuning, since we are utilizing the same pretrained model and a subset of the global fine-tuning data.

\paragraph{Global Fine-tuning} This approach entailed comprehensive fine-tuning of the pretrained ClinicLLM using all available History and Physical (H\&P) notes from the four [HOSPITAL] hospitals. 

\paragraph{Local Fine-tuning (Hospital Specific)} This approach involved the fine-tuning of the pretrained ClinicLLM model for each of the four [HOSPITAL] hospitals independently with the goal of adapting ClinicLLM to the unique characteristics of each hospital during the fine-tuning process, thereby enhancing the model's generalization. Performance was evaluated using testing datasets for each hospital separately. 

\begin{figure}[h]
  \centering
  \includegraphics[width=0.8\textwidth]{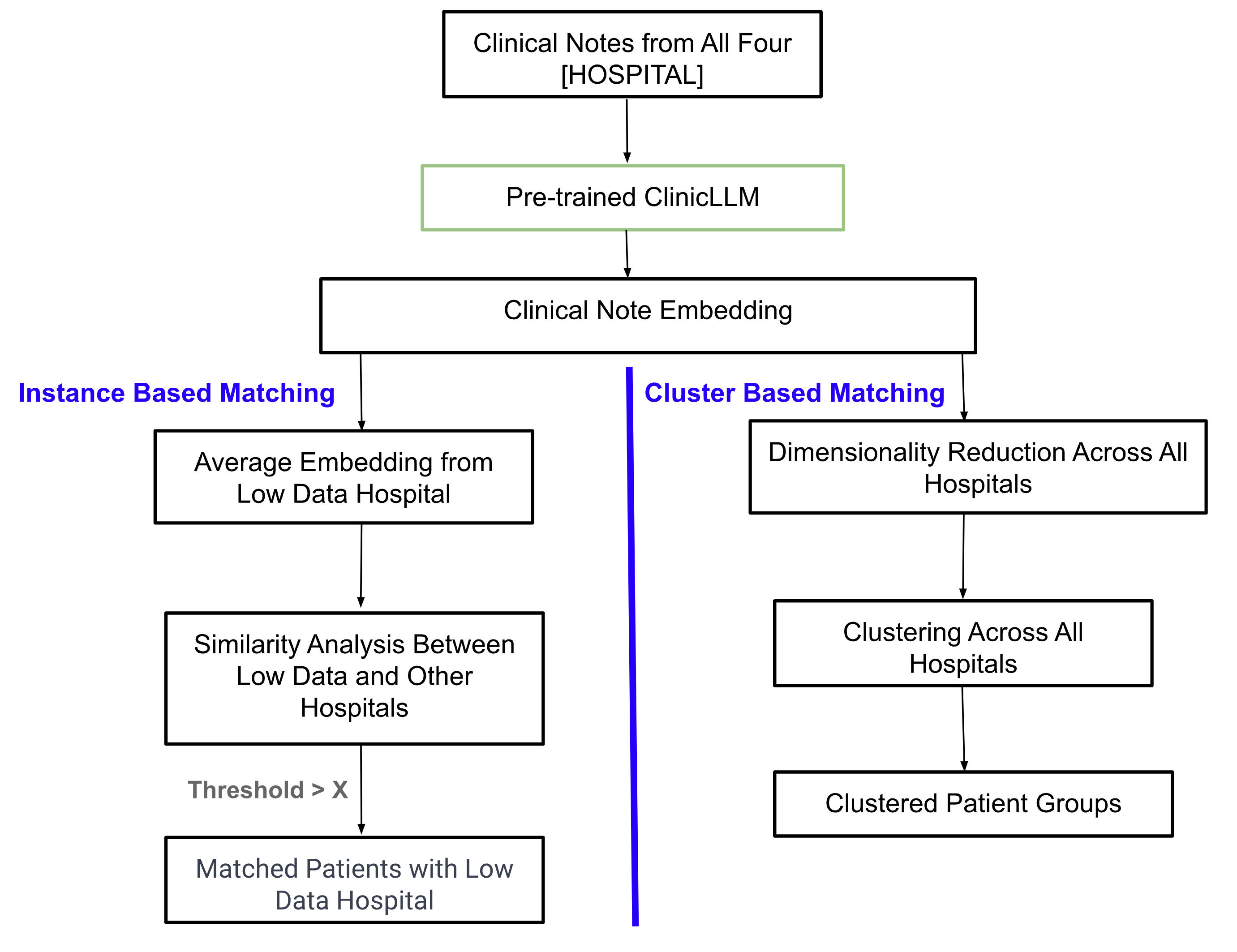}
  \caption{Workflow for both instance-based matching for identifying patient matches in low-data hospitals (Hospital 3 and Hospital 4) [left], and cluster-based matching to group patients with similar characteristics [right].}
  \label{fig:clustering-workflow}
\end{figure}

\paragraph{Instance-Based Augmented Fine-tuning} Instance-based augmented fine-tuning comprises a two-step process first identifying matched samples, followed by local fine-tuning on these samples (see Figure \ref{fig:clustering-workflow} for overview).  

For hospitals with the lowest number of fine-tuning samples which could potentially affect generalization of ClinicLLM (Hospital 3 and Hospital 4), we implement a method that augments data for hospitals with limited samples by finding similar patient embeddings from other hospitals \cite{bartolini2023data}. Initially, embeddings for all clinical notes are generated using the pretrained ClinicLLM. Then, mean of the token embeddings of each instance from Hospital 3 and Hospital 4 hospitals are calculated. Cosine similarity with embeddings from other hospitals is used to compare clinical notes irrespective of their length or scale. Matched samples are identified via the similarity exceeding a specific threshold. The threshold is defined via the average similarity of notes from the same patients across multiple visits. Then the mean of those average cosine similarity is computed, resulting in similarity thresholds of 0.94 and 0.92 for Hospital 3 and Hospital 4, respectively. This threshold definition was designed based on conversations with physicians at our institution; the intuition being that as we are considering only H\&P notes, there will likely be a significant overlap of information for the same patients across different encounters. 

Following identification of matched samples, local fine-tuning is conducted on these matched samples. Efficacy is assessed using AUC, AUPR, and ECE using the same initial testing datasets. It should be noted that we also assessed a range of thresholds. Decreasing the threshold, for example using a mean cosine similarity across all Hospital 4 notes (producing a threshold of 0.88) results in a sample size of 146,133 which is close to those in the global fine-tuning case of 161,773, thus not affecting the performance. Thresholds above the selected range were also examined. While extreme thresholds (i.e., above 0.98) select for notes which are very similar, this results in very small sample sizes such that no readmission cases (outcome =1) are included and the model cannot learn at all with respect to the positive class.

\paragraph{Cluster-Based Fine-tuning}

Cluster-based fine-tuning begins with defining the cluster, followed by local fine-tuning within each cluster (Figure \ref{fig:clustering-workflow}). 
The rationale is to enable ClinicLLM to effectively learn the representation of patients with similar characteristics, and fine-tuning on similar clusters allowing the model to learn the unique features of each cluster and better adaptation to diverse patient profiles. Initially, clinical text is transformed into embeddings using the pre-trained ClinicLLM model. Then, the dimensionality of these embeddings from the original dimension of 768 is reduced using Uniform Manifold Approximation and Projection (UMAP). UMAP is chosen for based on its ability to handle non-linear relationships within the embeddings \cite{mcinnes2018umap} compared to methods like Principal Component Analysis which primarily reduce dimensions by identifying and leveraging linear correlations in high-dimensional embedding spaces. Finally, k-means is used to group samples into similar clusters. The optimal number of clusters, k=4, is determined using the elbow method, which involves calculating the sum of squared errors (SSE) for each value of k.

After fine-tuning each cluster, we evaluate the performance for each hospital using the testing dataset and calculating the mean of the metrics weighted by sample size (number of notes from each cluster and hospital). 

\subsection{Population and Hospital Groups for Generalizability Assessment}
\label{method:rq1}

\paragraph{Hospital Group} Here the H\&P testing dataset is divided into four groups corresponding to the four [HOSPITAL] hospitals: Hospital 1, Hospital 2, Hospital 3, and Hospital 4. The model's performance in each specific hospital setting is evaluated. 

\paragraph{Insurance Type} The generalization of ClinicLLM is evaluated across the four main insurance groups:

\begin{itemize}

\item \textbf{Government Insurance}: This group includes Medicaid for low-income individuals and families, Medicare for those aged 65 and over and certain disabilities, Tricare for military personnel and their dependents, Champva for beneficiaries of the Department of Veterans Affairs, and Worker's Comp for work-related injuries.

\item \textbf{Private or Commercial Insurance}: This category encompasses health plans from private entities such as Blue Cross, other commercial insurance plans, and managed care arrangements like HMOs, Medicare HMOs, and PPOs.

\item \textbf{Self-Pay}: Patients in this category cover their medical expenses independently, without insurance coverage.

\item \textbf{Other Specific or Specialized Insurance}: This includes No Fault insurance, typically for auto incidents, and an 'Other' category for non-standard insurance types.
\end{itemize}

\paragraph{Race Category} Race groups were consolidated into six categories based on the CDC National Center for Health Statistics Race and Hispanic Origin Information guidelines (American Indian or Alaska Native, Asian, Black or African American, Native Hawaiian or Other Pacific Islander, and White)\footnote{https://www.cdc.gov/nchs/nhis/rhoi/rhoi\_glossary.htm} and one 'Unknown' category. 

\paragraph{Age Group} We used: Under 18, Young Adult (18-35), Adult (35-60), and Above 60.

\paragraph{Comorbidities} Comorbidity levels were calculated using the Charlson Comorbidity Index (CCI), which is derived from International Classification of Diseases (ICD) codes for chronic diseases based on the scoring function detailed in \citet{charlson1987new}. The comorbidity scores are categorized into four levels: Level 1 for low comorbidity (50th to 75th percentiles), Level 2 for moderate (75th to 90th percentiles), Level 3 for high (90th to 99th percentiles), and Level 4 for the most severe cases (above the 99th percentile).

\subsection{Metrics and Methods for Investigating the Generalization Gap} 
\label{method:rq2}

\paragraph{Metrics}

The model's performance and generalization to groups was assessed using: Area Under the Curve (AUC), Area Under the Precision-Recall Curve (AUPRC), and Expected Calibration Error (ECE). AUC was selected for its effective balance in evaluating false positives and negatives, crucial for accurate outcome prediction in cases such as 30-day readmission. AUC values typically range from 0.5, indicating no discriminative power, to 1.0, denoting perfect discrimination between classes. AUPRC was specifically chosen for scenarios with imbalanced datasets, where accurate identification of rare positive cases is essential. AUPR also ranges from 0 to 1, with higher values indicating greater precision and recall, thus reflecting better performance in identifying positive cases in imbalanced datasets \cite{saito2015precision, petersen2023assessing}. ECE values typically range from 0 to 1, with values closer to 0 indicating better calibration of the model. Lower ECE values are desirable as they imply that the predicted probabilities of outcomes are more accurately aligned with the actual outcomes \cite{guo2017calibration}.

\paragraph{Assessing Model Performance Under Varying Sample Size} 

An experiment was designed where equal numbers of samples were fine-tuned from both Hospital 1 and Hospital 2. We randomly selecting samples, repeated five times, for fine-tuning Hospital 1 and Hospital 2 and then calculating the mean AUC for each. 

\paragraph{Perplexity Analysis}

Perplexity, which indicates the predictability of text by a language model, was computed using a Masked Language Modeling (MLM) approach \cite{wang2019bert}. Perplexity is computed by individually masking each token in the input text, followed by the MLM predicting each masked token. Prediction accuracy is quantified using Cross-Entropy Loss for each token. The average loss across all tokens is computed and exponentiated to derive a perplexity score for each note. The median is chosen over measures like the mean to reduce the impact of outliers in the data. These scores range from 1 to infinity, with lower scores indicating better model performance, and higher scores suggesting that the model finds the text more unpredictable.

\paragraph{Clustering and Decision Tree} For fine-tuning and examining joint features that affect generalizability, patient notes are grouped into four clusters. This involves generating embeddings, reducing dimensions using UMAP, and applying K-means to determine notes group into the same cluster. After grouping notes into four clusters, nine features are used to describe these clusters: patient characteristics (comorbidity level, age, insurance type, race, borough), hospital data (readmission and mortality rate, hospital name), and note attributes (number of words in notes). A decision tree is used to classify these clusters based on these nine features. 

\subsection{Metrics for Improving Generalization}
\label{method:rq3}

For fine-tuning enhancement strategies, performance is evaluated on the test dataset. We also measure improvement using a proportional AUC change metric. For local, instance-based, and cluster-based fine-tuning, this is given by:
\begin{equation*}
    \text{Proportional AUC Change} = \frac{\text{AUC}_{\text{Local/Instance/Cluster}} - \text{AUC}_{\text{Base, Specific Hospital}}}{\text{AUC}_{\text{Base, Specific Hospital}}}
\end{equation*}
where \( \text{AUC}_{\text{Base, Specific Hospital}} \) is the AUC of the specific hospital on global fine-tuning.

For global fine-tuning, the proportional AUC change is:
\begin{equation*}
    \text{Proportional AUC Change} = \frac{\text{AUC}_{\text{Global}} - \text{AUC}_{\text{Base, Global}}}{\text{AUC}_{\text{Base, Global}}}
\end{equation*}
Here, \( \text{AUC}_{\text{Base, Global}} \) is the globally trained AUC encompassing the entire dataset, irrespective of any group.

\section{Generalization Across Population and Hospitals}
\label{results: rq-1}
The global baseline performance, denoted as $\text{AUC}_{\text{Base, Global}}$, is used to provide a reference point for comparing across different groups. On the random test set, the baseline performance metrics were as follows: $\text{AUC}_{\text{Base, Global}}$ = 76.90\%, AUPR = 34.40\%, and ECE = 0.20. For the temporal test set, $\text{AUC}_{\text{Base, Global}}$ was 73.60\%, AUPR 31.10\%, and ECE 0.22, indicating a decrease in performance on new types of data not previously encountered during training. The baseline $\text{AUC}_{\text{Base, Global}}$ indicates the AUC value across all the random or temporal testing datasets, irrespective of their group. In the following subsections, generalization performance to specific subgroups is reported. While the model's performance on both random and temporal test datasets was evaluated, we primarily focus on the temporal test results because it is a more challenging task, and more closely mimics the deployment scenario. 

\subsection{Generalization Across Hospitals}

Evaluating performance by hospital, we find that AUC values for Hospital 3 and Hospital 4 are 69.90\% and 51.20\% respectively. Compared to the temporal test set's global baseline AUC of 73.60\%, this represent a decrease of 3.70\% for Hospital 3 and 22.40\% for Hospital 4. Furthermore, when compared to the best-performing hospital (Hospital 1, with an AUC of 74.60\%), Hospital 3's AUC is 4.70\% lower and Hospital 4 is 23.40\% lower. (Table \ref{tab:global-temporal-metrics}). This indicates poorer generalization at the Hospital 3 and Hospital 4 relative to Hospital 1. The AUPR and ECE metrics also show poorer performance for Hospital 3 and Hospital 4 compared to Hospital 1. Specifically, Hospital 3's AUPR is 20.70\% and ECE is 0.27, versus Hospital 1's 34.10\% (AUPR) and 0.21 (ECE). Hospital 4's AUPR and ECE are 14.40\% and 0.42, respectively, also lagging behind Hospital 1. These metrics indicate worse prediction, discrimination and calibration for the two smaller hospitals. Detailed performance results by hospital are in Table \ref{tab:global-temporal-metrics}.

\begin{table}
    \centering
\caption{Performance on 30 day readmission task using global fine-tuning and temporal test data, across different hospitals and patient groups. The table also includes the readmission rate and the number of samples used during fine-tuning.}
\label{tab:global-temporal-metrics}   
\small
\begin{tabular}{llccccc}
        \toprule
        Group & Item & AUC (\%) & AUPR (\%) & ECE & Readmission Rate (\%) & Sample Size\\
        \midrule
        \multirow{4}{*}{Hospital} & Hospital 1 & 74.60 & 34.10 & 0.21 & 14.80 & 102,275 \\
                                 & Hospital 2 & 73.04 & 29.69 & 0.22 & 13.70 & 51,545 \\
                                  & Hospital 3 & 69.90 & 20.70 & 0.27 & 9.70 & 4,502 \\
                                  & Hospital 4 & 51.20 & 14.40 & 0.42 & 14.40 & 3,451 \\
        \midrule
        \multirow{4}{*}{Insurance Type} & Government & 65.15 & 32.72 & 0.22 & 20.30 & 54,705 \\
                                 & Private & 76.43 & 30.01 & 0.22 & 11.20 & 105,328\\
                                  & Self-Pay & 77.78 & 13.02 & 0.38 & 6.30 & 1,257\\
                                  & Other & 64.03 & 16.71 & 0.35 & 16.80 & 483\\
        \midrule
        \multirow{6}{*}{Race Group} & White & 72.68 & 30.06 & 0.22 & 14.40 & 89,273 \\
                                 & Black & 71.71 & 33.10 & 0.21 & 15.80 & 19,207 \\
                                  & Asian & 76.56 & 33.84 & 0.19 & 23.00 & 16,592 \\
                                  & {\shortstack[l]{American Indian\\ or Alaska Native}} & 81.27 & 34.03 & 0.24 & 7.20 & 1,068 \\
                                  & {\shortstack[l]{Native Hawaiian\\ or Other Pacific Islander}} & 57.82 & 8.96 & 0.42 & 9.20 & 704 \\
                                  & Unknown & 75.10 & 31.23 & 0.22 & 14.00 & 34,929 \\
        \midrule
        \multirow{4}{*}{Age Group} & Under 18 & 75.21 & 26.53 & 0.21 & 4.50 & 24,147\\
                                 & Young Adult (18-35) & 80.81 & 23.88 & 0.26 & 8.50 & 16,707\\
                                  & Adult (35-60) & 74.69 & 31.07 & 0.21 & 11.30 & 40,937\\
                                  & Above 60 & 64.75 & 32.06 & 0.22 & 20.00 & 79,858 \\
        \midrule
        \multirow{4}{*}{Comorbidities} & Level 1 (Low) & 74.76 & 24.46 & 0.25 & 9.40 & 110,258\\
                                 & Level 2 (Moderate) & 66.86 & 33.69 & 0.22 & 20.30 & 218,30\\
                                  & Level 3 (High) & 61.43 & 37.93 & 0.20 & 27.10 & 251,60\\
                                  & Level 4 (Severe) & 58.08 & 43.25 & 0.19 & 33.00 & 4,525\\
        \bottomrule
    \end{tabular}
\end{table}

\subsection{Generalization Across Patient Groups by Insurance Type}

The model shows reduced effectiveness distinguishing 30-day readmission likelihood in notes from patients with government insurance (AUC: 65.15\%) and those in the `Other'/unspecified insurance category (64.03\%) compared to the global baseline AUC of 73.60\%. AUC values for the government and other/unspecified insurance categories are 12.63\% and 13.75\% lower, respectively, compared to those with Self-Pay (AUC: 77.78\%). Though notes from the Self-Pay patient insurance category had a higher AUC (77.78\%); the AUPR (13.02\%) and ECE (0.38) are comparatively poorer and the sample size in the Other (n=483) and Self-Pay (n=1,257) groups are fairly small (Table \ref{tab:global-temporal-metrics}).

\subsection{Generalization Across Patients by Race Category}

The model shows best performance for notes from patients from the American Indian (AUC: 81.27\%) and Asian (AUC: 76.56\%) categories, surpassing the global baseline AUC of 73.60\%. The Asian group has a relatively high number of notes (n=16,592), and a higher rate of readmission (23.00\%) as shown in Table \ref{tab:global-temporal-metrics}, while the American Indian group had a small sample size (n=1,068). The Native Hawaiian or Other Pacific Islander category had the lowest AUC, and only had a sample size of n=704 for the model to learn from. The model performs slightly worse in notes from patients in the Black racial group compared to the white group (71.71\% vs. 72.68\%).

\subsection{Generalization Across Patients by Age Group}

The model encounters the greatest challenges in generalizing for the age group Above 60, with an AUC values of 64.75\%, which is 8.85\% lower compared to the global baseline AUC of 73.60\%. This is despite this group including the highest numbers of patients and notes (n=79,858), as well as a higher readmission rate (20.00\%) and mortality (3.40\%). The AUC values for this group are 16.06\% lower compared to best-performing Young Adult group (AUC: 80.81\%).

\subsection{Generalization Across Patients by Comorbidities}

Compared to the global baseline AUC of 73.60\%, ClinicLLM faces challenges in distinguishing positive and negative readmission in notes from patients with higher levels of comorbidities. Specifically, for patients with comorbidity Level 3 and Level 4, the AUC values are 61.43\% and 58.08\% respectively, which are 12.17\% and 15.52\% lower than the global baseline (Table \ref{tab:global-temporal-metrics}). The AUC performance for Level 3 and 4 is also 13.33\% and 16.68\% lower for these groups, respectively, compared to patients with low comorbidities (Level 1) who have an AUC of 74.76\%. Interestingly, patients in these higher comorbidity groups, including Level 2, exhibit a higher AUPR of 33.69\%, 37.93\%, and 43.25\% for Level 2, 3, and 4 respectively. This suggests that while the model struggles to differentiate between positive and negative readmission cases, it performs comparatively better in predicting true positive readmission cases than in patients with low comorbidity (Level 1). Level 1 has lower readmission rates (9.40\%) thus the model had fewer readmission samples for learning to accurately predict positive readmission cases in patients with low comorbidity (Table \ref{tab:global-temporal-metrics}).

\section{Investigating Reasons for Lack of Generalization}
\label{results: rq-2}

\subsection{Levels of Analysis}

Here we assess factors affecting the generalization capabilities of clinical large language models (LLMs) by examining attributes at hospital, patient, and note levels. 

\begin{figure}[h]
  \centering
  \includegraphics[width=0.8\textwidth]{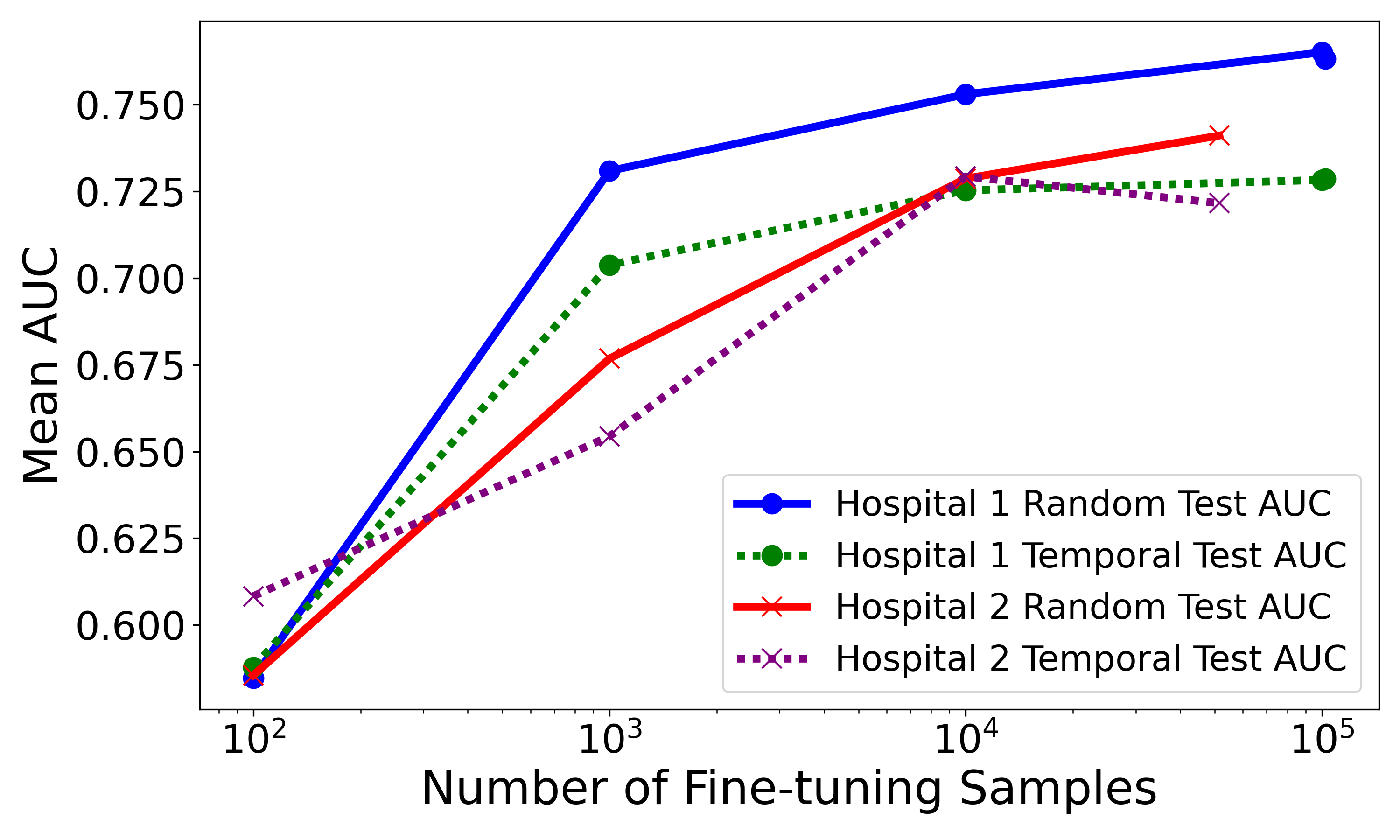}
  \caption{Comparative AUC analysis for Hospital 1 and Hospital 2 in random and temporal tests, distinguished by solid and dotted lines respectively, demonstrates that with an equal number of fine-tuning samples, Hospital 1 generalizes better than Hospital 2. The analysis was conducted five times for each sample size using a random seed. The standard deviation for each sample size was almost zero, except for standard deviations of $0.02$ for the Hospital 1 random and temporal tests, both with a sample size of $10^2$.}
  \label{fig:sample_size_vs_metrics}
\end{figure}

\paragraph{Impact of Sample Size on Hospital-Level Model Performance}

At the hospital level, a majority of the data is sourced from Hospital 1 (along with a higher comparative readmission rate) enabling the model to learn more effectively about patients from this hospital. This could potentially lead to better model generalization for this specific hospital compared to others. In analysis of sample size versus performance for the Hospital 1 and Hospital 2, we found that the mean AUC for Hospital 1 in both random and temporal tests, and for Hospital 2 in the random test, consistently improves with an increasing number of fine-tuning samples (Figure \ref{fig:sample_size_vs_metrics}). A significant phase in learning is identified when the sample size increases from 100 to 1,000, marked by substantial improvement in AUC for both hospitals (all curves start to improve more slowly after 10,000 samples). This indicates that this sample size range is crucial for the model’s learning and generalization capabilities, emphasizing the importance of sample size in model performance. As expected, the random test results in higher performance for both hospitals given the added information it contains. Yet, with an equal number of samples, both for the random and temporal test setups, Figure \ref{fig:sample_size_vs_metrics} reveals that performance disparities between Hospital 1 and Hospital 2 persist. This suggests that sample size alone does not fully account for model generalization. Other factors such as patient health status, comorbidity, and age, or note-level factors could also play a role in model generalizability.


\paragraph{Perplexity Analysis of Language Models across Hospitals}

In the random test, the median perplexities for Hospital 1, Hospital 2, Hospital 3, and Hospital 4 are 2.29, 2.41, 2.32, and 2.14, respectively. In the temporal test, these values are 3.07, 3.01, 2.97, and 3.01, respectively. Perplexity shows consistency across all hospitals, with marginally better results in the random test dataset than in the temporal dataset. The elevated perplexity in the temporal test set, compared to the random test set, aligns with expectations, as it includes new patients whose information was not present during the pre-training phase. Overall this evaluation indicates consistent predictive power across all hospitals.

\subsection{Cluster Analysis}
\label{results:rq2-cluster}
\begin{figure}[h]
  \centering
  \includegraphics[width=0.9\linewidth]{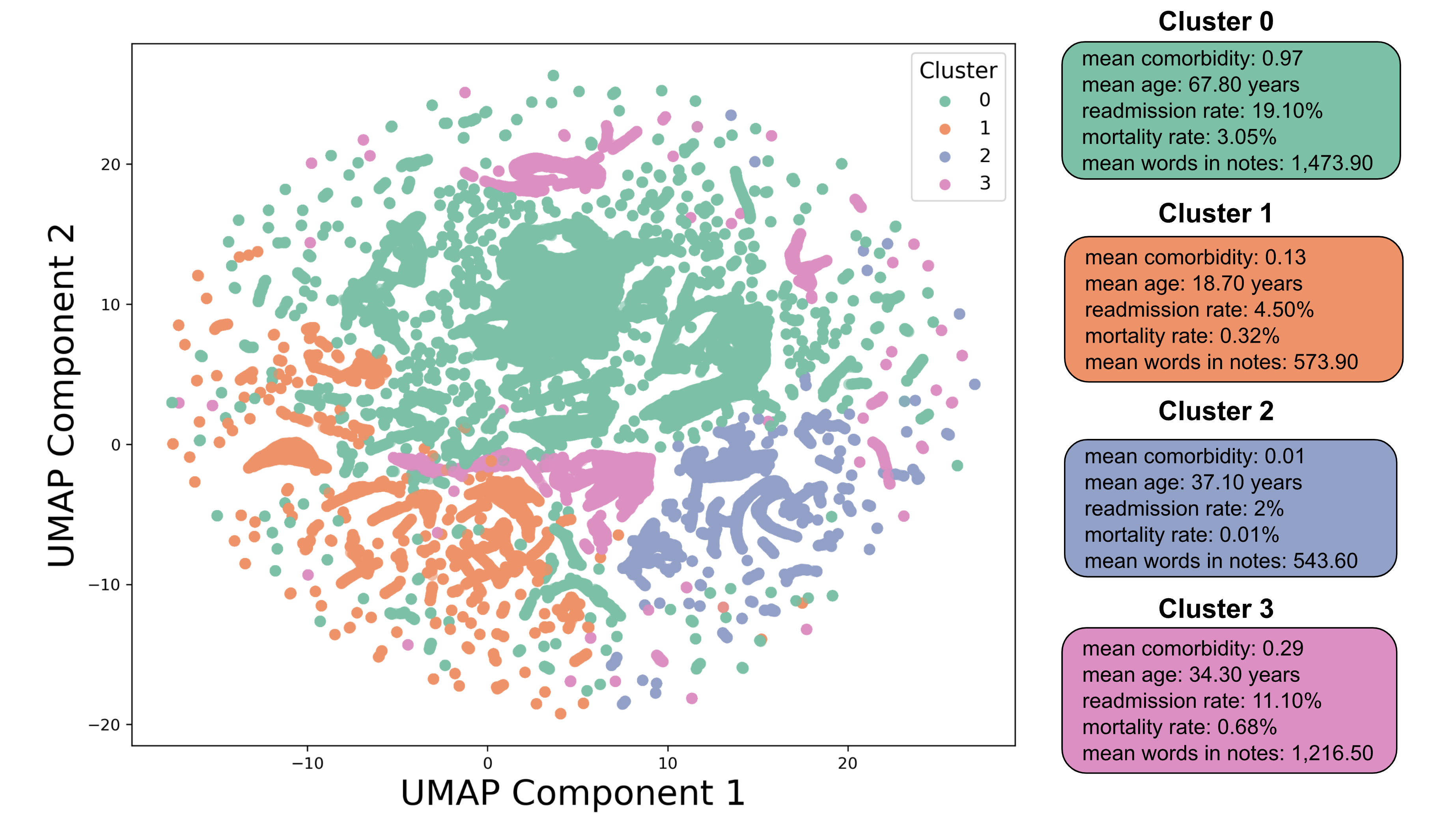}
  \caption{Visualization and report of important factors distinguishing clusters of similar notes via patient, hospital, and note-level characteristics. The distinguishing factors include hospital-level features like readmission and mortality rates, as well as patient and note-level features: mean comorbidities, age of patient, and number of words in the notes (the top three features in the decision tree).}
  \label{fig:cluster-visualization}
\end{figure}

Clustering similar notes shows four qualitatively different groups of patients. Cluster 0 mainly includes elderly patients with high comorbidity levels and lengthy notes. Cluster 1 is comprised of younger patients who are healthier with notes of moderate length. Cluster 2 consists largely of adult patients who are generally healthy and includes the shortest average note length. Cluster 3 includes adult patients with detailed notes (Figure \ref{fig:cluster-visualization}). It should be noted that the clusters were fairly evenly distributed across hospitals (e.g. Clusters 0-3 with 59.99\%, 67.51\%, 69.47\% and 58.67\% from Hospital 1 which overall has 63.32\% of the sample size); not skewed towards a particular hospital, indicating representation across all the [HOSPITAL] hospitals. From the decision tree plot and analysis of the clusters, the top three features contributing to differentiation of clusters were: the age of the patients, number of comorbidities, and the number of words in the notes.

\section{Strategies for Improving Generalization}
\label{results: rq-3}

Baseline performance for each hospital (to compare with local, augmented and cluster-based fine-tuning approaches) is from the global fine-tuning AUC metric on the temporal test set for that hospital. For Hospital 1 is 74.60\%, Hospital 2 it is 73.04\%,  Hospital 3 it is 69.90\%, and for Hospital 4 it is 51.20\%. 

\subsection{Local Fine-tuning}

To assess strategies for improving generalization of clinical LLMs, we first conducted fine-tuning of ClinicLLM via fine-tuning specific to each of the four different [HOSPITAL]. In temporal tests, we noted an enhancement in generalization for Hospital 2 and Hospital 3, with an AUC increase from the base level of 73.04\% to 73.22\% and 69.90\% to 71.57\% respectively. For Hospital 4, the performance improved from a base level random guessing (AUC 51.20\%) to 57.21\%, as shown in Table \ref{tab:rq-3-result-finetuning}. The proportional AUC change in Table \ref{tab:rq-3-result-finetuning} for Hospital 1, Hospital 2, Hospital 3 and Hospital 4 are -0.87\%, 0.25\%, 2.39\%, and 11.74\% respectively, showing improvement on Hospital 2, Hospital 3, and Hospital 4. Additionally, local fine-tuning resulted in better calibration than global fine-tuning across all hospitals, as indicated by lower Expected Calibration Error (ECE). While the performance on Hospital 1 remained nearly the same in both local and global fine-tuning, local fine-tuning also has distinct advantages, such as faster computation and reduced data requirements.

\begin{table}
    \centering
\caption{Predictive performance on the 30-day readmission task on a temporal test set using: 1) global fine-tuning on the entire dataset, 2) local fine-tuning on hospital-specific datasets, 3) instance-based augmented fine-tuning on a matched dataset for a low data hospital, and 4) cluster-based fine-tuning. For global fine-tuning, the proportional AUC change is measured relative to the base AUC on the entire dataset (73.60\%). For local, instance-based augmented, and cluster-based fine-tuning, the proportional AUC change is measured relative to the corresponding hospital AUC in global fine-tuning, to assess the change in each specific hospital. In cluster-based fine-tuning, the weighted mean of the cluster metrics is computed using the sample sizes of notes for each hospital. The sample size in the table indicates the number of samples used during the fine-tuning processes.}
\label{tab:rq-3-result-finetuning}   
\small
\begin{tabular}{lccccc}
        \toprule
        Types of Fine-tuning & Hospital/Sample Size & AUC (\%) & AUPR (\%) & ECE  & Proportional AUC Change (\%)\\
        \midrule
        \multirow{4}{*}{Global Fine-tuning} & Hospital 1/161,773  & 74.60 & 34.10 & 0.21 & 1.36  \\
                                 & Hospital 2/161,773 & 73.04 & 29.69 & 0.22 & -0.76  \\
                                  & Hospital 3/161,773 & 69.90 & 20.70 & 0.27 & -5.03  \\
                                  & Hospital 4/161,773 & 51.20 & 14.40 & 0.42 & -30.43 \\
        \midrule
        \multirow{4}{*}{Local Fine-tuning} & Hospital 1/102,275 & 73.95 & 34.10 & 0.12 & -0.87 \\
                                 & Hospital 2/51,545 & 73.22  & 29.02 & 0.22 & 0.25 \\
                                  & Hospital 3/4,502 & 71.57 & 19.86 & 0.08 & 2.39 \\
                                  & Hospital 4/3,451 & 57.21 & 15.57 & 0.22 & 11.74 \\
        \midrule
        \multirow{2}{*}{\shortstack[l]{Instance-Based \\  Augmented Fine-tuning}}
                                  & Hospital 3/103,657 & 67.90 & 20.73 & 0.28 & -2.86 \\
                                  & Hospital 4/107,286 & 49.98 & 16.54 & 0.44 & -2.38 \\
        \midrule
       \multirow{4}{*}{Cluster-Based Fine-tuning} & Hospital 1/102,275 & 69.99 & 28.28 & 0.26 & -6.18\\
                                 & Hospital 2/51,545 & 66.52 & 25.49 & 0.26 & -8.93\\
                                  & Hospital 3/4,502 & 70.27 & 20.00 & 0.31 & 0.53 \\
                                 & Hospital 4/3,451 & 50.54 & 24.83 & 0.38 & -1.30 \\
        \bottomrule
    \end{tabular}
\end{table}

\subsection{Fine-tuning with Augmented Data}

Instance-based augmented fine-tuning was applied to hospitals with lower data availability (Hospital 3 and Hospital 4), following the method described in Section \ref{subsec: fine-tuning}. Fine-tuning ClinicLLM with augmented data for these hospitals showed a performance boost in the random test set. Notably, for Hospital 3, the AUC improved from base level of 71.20\% to 75.10\%. At Hospital 4, there was an increase from base of 70.90\% to 74.82\% in random test. However, in the temporal test, the proportional AUC decreased by -2.86\% and -2.38\% for Hospital 3 and Hospital 4, respectively as shown in Table \ref{tab:rq-3-result-finetuning} which focuses on the temporal test results (not random test results) due to their relevance and added difficulty.


\subsection{Cluster-based Tuning}
\label{result3-cluster-finetuning}

Table \ref{tab:rq-3-result-finetuning} indicates the weighted sample size mean of the AUC, AUPR, and ECE metrics, calculated based on the sample sizes across different cluster-based fine-tuning methods. Only Hospital 3 showed an improvement, with an AUC of 70.27\%, which is a 0.53\% increase compared to the baseline global fine-tuning AUC of 69.90\% for Hospital 3. Hospital 2 experienced the largest decrease, with an AUC of 66.52\% compared to the global fine-tuning baseline AUC of 73.04\% for Hospital 2, a reduction of -8.93\%. Hospital 1 and Hospital 4 also showed reductions of -6.18\% and -1.30\%, respectively, compared to their global fine-tuning baselines of 74.60\% and 51.20\%. Additionally, compared to the global fine-tuning baseline, the AUPR for Hospital 1 decreased from 34.10\% to 28.28\%, and for Hospital 2, it decreased from 29.69\% to 25.49\% in cluster-based fine-tuning. This indicates that for these two hospitals, cluster-based fine-tuning struggled to accurately predict true positive cases compared to the global fine-tuning baseline.

\section{Discussion and Conclusion}

Our findings provide several insights regarding the generalizability of large language models in clinical systems. First, we show that the ClinicLLM model had the poorest performance in hospitals with lower numbers of note samples, specifically the Hospital 3 and Hospital 4. These hospitals had the lowest AUC, AUPR, and the highest ECE compared to the larger Hospital 1 and Hospital 2. Additionally, the model worked worst for patients covered by government and unspecified insurances compared to patients having private or self-pay as they also had the lowest performance. Similarly, the lowest AUC resulted for elderly patients (over the age of 60), and those with high comorbidities compared to their respective race, age and comorbidity groups. Within race group categories, poorer generalization was observed for patients in the Native Hawaiian or Other Pacific Islander group (very low sample size) and Black racial group (slightly lower than for those in the white group).

We found that sample size alone will not improve generalizability. This was evidenced through hospital level analysis, where we observed that with the increasing number of fine-tuning samples the AUC increases which indicates that sample size plays a role in model performance. Despite these trends, with the same number of samples, AUC in different hospitals can still vary. Multiple factors contribute to generalizability: Based on a supervised learning analysis, the top three features that distinguish generalizability performance were: health status, age, and number of words in note.

Our results showed that hospital specific local fine-tuning helps improves generalization, even within settings with limited number of notes. This was evidenced by observing improved AUC for Hospital 2 (0.25\%; n=51,545), Hospital 3 (2.39\%; n=4,502), and Hospital 4 (11.74\%, n=3,451) (all hospitals except Hospital 1, which has the largest set of notes) compared to global performance on the model fine-tuned on all notes. On the other hand, instance-based augmented fine-tuning for low sample size between notes helps patients coming from the same timeframe as the pre training and fine-tuning (random test sets), but it does not help generalization for patients in a new timeframe (temporal test sets). (improvements in the AUC for Hospital 3 and Hospital 4 in random tests (5.48\% and 5.53\% respectively) but no improvement in temporal testing sets (-2.86\% and -2.38\%)). Additionally, we observed that fine-tuning with similar note clusters did not improve performance, evidenced by negative proportional AUC changes across hospitals when compared to base-line performance on a globally fine-tuned model, except for Hospital 3 hospital (which had the relatively highest proportion of notes from Clusters 1 and 2 which were low comorbidity, low mortality rate, low mean age, and potentially easier to predict from). These findings, including the lack of improvement in AUC on temporal test sets, and instance or cluster based fine-tuning, alongside improvements with local hospital specific fine-tuning, suggest that fine-tuning with specific hospital notes helps due to unique characteristics that may not be captured in the attributes considered. For example, variability in hospital systems, specializations, and types of patients may be relevant when predicting readmission, especially when tested on new patients. 

Previous studies have demonstrated the effectiveness of clinical large language models (LLMs) in clinical prediction tasks \cite{jiang2023health, singhal2022large, yang2022large}. Our research extends their work by identifying key generalization factors such as sample size, health status, and patient age, and explore local, instance-based augmented, and cluster-based fine-tuning strategies for improving the generalization capability of LLMs across different hospitals and patient groups. 

Relevant to practitioners and those designing LLMs for healthcare systems, our study finds poorer generalization in samples with lower number of notes, patients with government and unspecified insurance, patients from the Black racial group, older patients, and those with severe health conditions. This highlights specific groups which practitioners may take a closer look at generalization for before deploying clinical Large Language Models (LLMs) even within the same healthcare systems. At the same time for the design of such systems, it should be noted that it is not just one factor that will be relevant; our study also shows that the confluence of multiple factors, including sample size, health status, and patient age combined, affect the generalization of clinical LLMs. Finally, our study provides evidence that local fine-tuning with a small subset of patients from the same hospital aids in generalization compared to augmenting data from other hospitals. This result, along with observations related to differential performance even at similar sample sizes shows that there may be hospital-specific factors in addition to the ones considered here, which are related to prediction from notes. Researchers who are trying to improve generalization through data augmentation and patient similarity should consider this important finding. 

There are a few limitations to note from this work. In our fine-tuning process, the focus was exclusively on History and Physical (H\&P) notes from the four major [HOSPITAL] hospitals. This creates a gap in determining whether our findings on generalization are applicable to, for example, other [HOSPITAL] outpatient facilities, as clinical data from those hospitals may differ in form and content from the data of the four hospitals we analyzed. Additionally, we utilized a fixed pre-trained model without engaging in-hospital pre-training or adversarial training. Previous research has indicated that such strategies could enhance generalization \cite{dong2021should, hendrycks2020pretrained}, but they were not used in our study due to their high costs and computational demands. Furthermore, in our cluster-based fine-tuning, we focused on only four clusters due to the scalability of fine-tuning by clusters. Our findings indicate that future work should extend the analysis to more granular or precise measures of patient similarity for clustering to improve performance. 

Lastly, our model follows the original BERT architecture, which is limited to processing a maximum of 512 tokens, a limit set by considerations of effectiveness and computational constraints. This results in the truncation of clinical notes and the potential loss of detailed information. Future studies could explore methods to handle longer clinical notes \cite{yang2020beyond} and assess their impact on the generalization abilities of clinical LLMs.

\section*{Funding/Support}
We are grateful for support for this research from Optum, and from the National Science Foundation (grant number 1845487).

\bibliographystyle{ACM-Reference-Format}
\bibliography{sample-base}

\end{document}